\documentclass[letterpaper, 10 pt, conference]{ieeeconf}  

\IEEEoverridecommandlockouts                              

\usepackage{amssymb}  
\usepackage{amsmath}
\usepackage{physics}
\usepackage{cite}
\usepackage{subfigure}
\usepackage{graphicx}
\usepackage{subcaption}
\usepackage{bm}
\usepackage{lipsum}
\usepackage{multirow}
\usepackage{graphics}
\usepackage{xcolor}
\usepackage{hhline}
\usepackage{hyperref}
\usepackage{siunitx}
\usepackage{mathtools, cuted}
\usepackage{tikz}         
\usepackage{algorithm}    
\usepackage{algorithmic}  

\renewcommand{\baselinestretch}{0.98}

\begin{document}

\title{Geometry-Aware Visual Odometry for Bronchoscopic Navigation via High-Gain Observer Fusion}

\author{
Mohammadreza Kasaei$^{1,2}$,
Francis Xiatian Zhang$^{2}$,
Feng Li$^{2}$,
Farshid Alambeigi$^{3}$, \\
Kevin Dhaliwal$^{2}$,
Mohsen Khadem$^{1,2}$%
\thanks{$^{1}$School of Informatics, University of Edinburgh, UK.}%
\thanks{$^{2}$Baillie Gifford Pandemic Science Hub, Institute for Regeneration and Repair, University of Edinburgh, UK.}
\thanks{$^{3}$Walker Department of Mechanical Engineering and Texas Robotics, The University of Texas at Austin, USA.}%
}
\maketitle

\thispagestyle{empty}
\pagestyle{empty}

\begin{abstract}
Navigational bronchoscopy is critical for pulmonary interventions, yet current platforms depend heavily on pre-operative CT or external sensors, limiting their use in critical care and resource-constrained settings. Vision-only navigation offers a scalable alternative, but conventional visual odometry (VO) struggles with texture-poor airway images, specularities, and the vanishing-point singularities of tubular anatomy, leading to frequent tracking failures and drift.  
We present a geometry-aware VO framework that explicitly leverages vanishing-point cues from airway lumens. Detected lumens are back-projected to 3D rays, whose weighted fusion yields a stable forward heading even when parallax cues are absent. This heading, together with looming-based velocity estimates, is fused with noisy VO outputs using a bespoke high-gain observer that enforces airway-following priors and rejects drift.  
We validate the method on ex-vivo mechanically ventilated human lungs with electromagnetic tracking ground truth. Compared to state-of-the-art pipelines (ORB-SLAM2, LoFTR-VO, DPVO), our approach reduces absolute trajectory error by more than 50\% and achieves the lowest relative pose error across all test sequences. 
\end{abstract}

\section{INTRODUCTION}

Navigational bronchoscopy is widely used for lung cancer diagnosis and pulmonary interventions. Commercial systems range from manual platforms such as superDimension (Medtronic) \cite{Rivera2013} to robotic solutions including Ion (Intuitive Surgical) \cite{REISENAUER2022308}, Monarch (J\&J) \cite{rojas-solano}, and Galaxy (Noah Medical) \cite{saghaie}. These platforms integrate electromagnetic (EM) \cite{Khandhar2017} or optical shape sensing \cite{REISENAUER2022308} with pre-operative CT to register the bronchoscope within a 3D airway map, enabling precise guidance while reducing risks such as pneumothorax and hemorrhage \cite{Ricketse000595}. Despite their success, sensor-based systems require specialised infrastructure (e.g., EM tracking, fluoroscopy), while CT-guided navigation is unsuitable for many critical care patients where high-quality scans are not feasible \cite{Patolia2021BronchICU,Roder2024RespInfections}. Consequently, NB is largely limited to elective cancer diagnostics. Moreover, CT-based registration is error-prone due to respiratory motion, with bronchial tree shifting up to 25~mm per cycle \cite{breath_cycle}. 

An alternative is Virtual Bronchoscopy (VB), where CT-derived virtual images are matched to live video \cite{tian2025harnessing,banach2021visually}. Although commercial VB exists, systems are semi-automatic, require operator input, and degrade under axial rotations \cite{asano}. Improvements include handcrafted registration \cite{merritt2013,luo2012,shen2017}, deep learning \cite{shen2019,sganga2019autonomous}, and more recently view-synthesis with NeRF, which aligns real video to CT-trained radiance fields \cite{nerf}. Tian et al.~\cite{Tian2024} further proposed PANS, combining monocular depth-based motion inference with CT-informed airway semantics in a Monte Carlo framework for real-time robust localisation. Nevertheless, all CT-dependent methods remain constrained by the limitations of pre-operative imaging, especially in dynamic or intensive-care settings.

\begin{figure}[t]
    \centering
        \includegraphics[width=0.85\columnwidth]{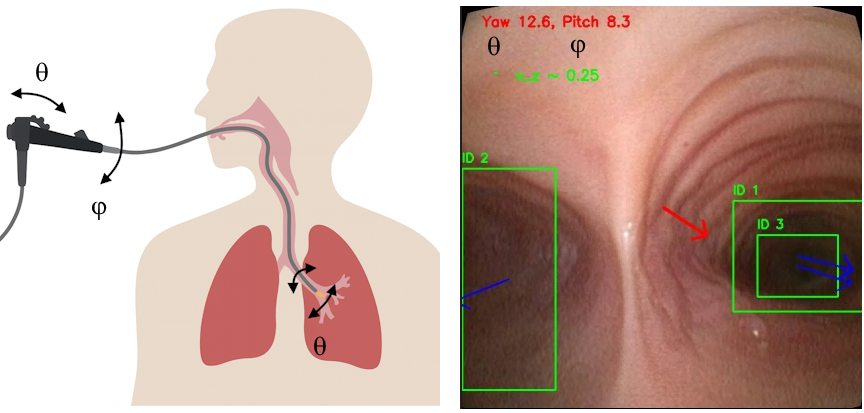}
    \caption{Vanishing-point geometry in bronchoscopy. The bronchoscope tip is steered by twist $\phi$ and bend $\theta$, but in tubular airways these motions produce little parallax and camera singularities that make direct orientation recovery unreliable. Distal lumen entrances, however, act as portals aligned with the true bronchial axis. By back-projecting their centers to 3D rays (blue arrows) and averaging them, we obtain a stable forward heading (red arrow) that serves as a geometry-aware orientation prior even when classical VO cues are weak or absent.}
    \label{fig:bronch}
\end{figure}

Vision-based methods offer a compelling alternative to CT or sensor-based navigation. By relying solely on sequential bronchoscopic images, Visual Odometry (VO) and Simultaneous Localization and Mapping (SLAM) aim to continuously track the bronchoscope’s camera pose and reconstruct an airway map in real time, thereby enabling repeated access to specific lung regions even in mechanically ventilated patients. VO/SLAM has been widely studied in other endoscopic domains such as colonoscopy, sinus endoscopy, and laparoscopy \cite{Liu2022,Mountney2010, GomezRodriguez2021}, motivated by the need to enhance intraoperative navigation, improve situational awareness, and increase the safety and efficacy of minimally invasive procedures. Applying VO/SLAM to bronchoscopy, however, presents unique challenges: bronchoscopic videos are texture-poor and repetitive, providing few stable features for classical keypoint-based methods such as SIFT \cite{sift} or ORB \cite{orb}, which often leads to frequent tracking failures \cite{odometry,deng_feature_bronch_2023}; lighting conditions further complicate tracking due to specular reflections, motion blur, and strong illumination changes from the endoscope light source. These factors make VO/SLAM in bronchoscopy an open research problem that continues to motivate ongoing investigation \cite{odometry}.

Recent research has made notable progress in addressing these challenges. 
Wang et al.~\cite{Wang2020} demonstrated a modified SLAM pipeline for bronchoscopy that improved feature matching robustness and achieved millimeter-level accuracy on phantom data. 
Deng et al.~\cite{deng_feature_bronch_2023} benchmarked classical versus learned feature detectors, showing that deep matchers such as LoFTR~\cite{sun2021loftr} significantly outperform traditional methods for intra-bronchial VO. 
In parallel, dense and learning-based VO/SLAM approaches have been developed to overcome scale ambiguities and texture limitations. 
Ozyoruk et al.~\cite{OZYORUK2021102058} introduced \textit{EndoSLAM}, which combines monocular depth and pose learning, and released a public endoscopic dataset including phantoms with ground-truth trajectories. 
Kalia et al.~\cite{Kalia2025} further advanced the field by introducing a self-supervised direct pose estimation framework that bypasses explicit depth reconstruction, instead leveraging temporal consistency and photometric alignment to predict bronchoscope motion directly from monocular images. 
However, the majority of these techniques have been evaluated only on short ($\sim$10~sec) sequences~\cite{deng_feature_bronch_2023} or synthetic datasets ~\cite{ Kalia2025, Wang2020}. 
Together, these efforts illustrate the promise of vision-only navigation while underscoring the need for solutions tailored to the unique challenges of bronchoscopy.

\subsection{Problem Statement and Contributions}

Visual Odometry (VO) forms the computational backbone of navigational bronchoscopy. When combined with pre-operative CT, it can mitigate drift and enable real-time registration of bronchoscopic video to a segmented airway tree; when used alone, VO provides a pathway to navigation in resource-constrained settings where CT is not feasible. This capability is critical for repeatable access to target sites during ICU procedures such as bronchoalveolar lavage (BAL), endobronchial or transbronchial biopsy, and targeted antimicrobial or chemotherapeutic delivery. Yet, applying VO in bronchoscopy is particularly challenging: airway images are texture-poor and lack salient features, lighting is unstable with specular reflections, intensity fluctuations, and motion blur, and the tubular geometry of bronchi introduces vanishing-point effects that collapse depth cues and render certain camera motions unobservable, making pose estimation fragile.

To address these challenges, we propose a novel VO framework that explicitly leverages vanishing-point geometry. First, we exploit the vanishing point to directly estimate camera motions that are typically unobservable in tubular environments, thus recovering information lost to classical VO pipelines (Fig.~\ref{fig:bronch}). These estimates are then combined with a bespoke observer that incorporates a geometric model of the airway tree. The observer uses both the vanishing-point estimates and the noisy, unreliable outputs of classical VO as priors, fusing them into a robust estimate of camera motion even in the presence of poor texture, specularities, and geometric singularities. 

We validate this approach on an ex-vivo mechanically ventilated human lung model, demonstrating reliable pose estimation in challenging regions of the bronchial tree where existing VO pipelines fail. To the best of our knowledge, this is the first VO framework tailored to the singularity structure of bronchoscopic navigation, providing a foundation for navigational bronchoscopy in critical care.

\noindent Our key contributions are:
\begin{itemize}
    \item A geometry-aware heading estimation method based on lumen-derived vanishing points, providing orientation cues in airway regions where parallax is weak.  
    \item A high-gain observer that fuses vanishing-point orientation, looming-based velocity, and noisy VO into a smooth and airway-consistent trajectory estimate.  
\end{itemize}

\section{Methodology}

\begin{figure}[t]
    \centering
        \includegraphics[width=1\columnwidth]{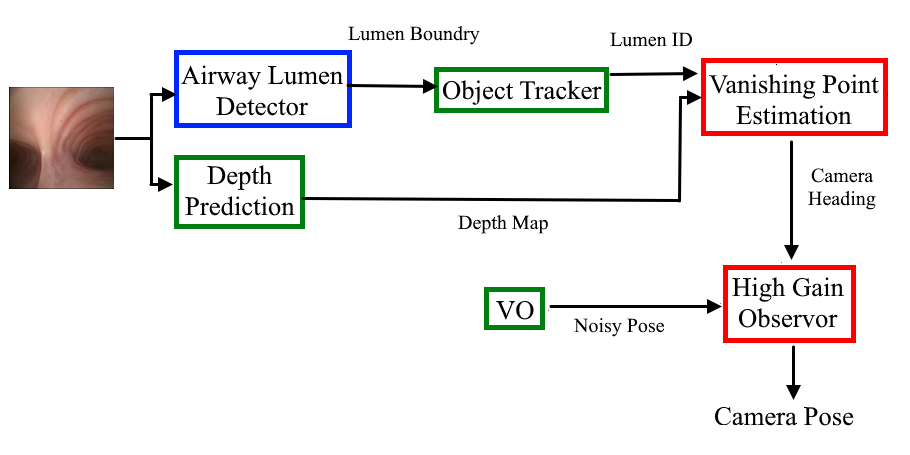}
    \caption{Block diagram of the proposed framework. Input bronchoscopic images are processed by a YOLO-based lumen detector and SORT tracker, while a parallel depth network refines lumen centers. These outputs are fused to estimate camera heading via vanishing-point geometry. The heading estimate is then combined with noisy VO in a bespoke high-gain observer, yielding a geometry-consistent and drift-robust bronchoscope pose. Green blocks denote off-the-shelf components, blue blocks fine-tuned modules, and red blocks bespoke designs.}
    \label{fig:block}
\end{figure}

Our proposed framework integrates geometric modeling, learning-based perception, and observer design to achieve robust vision-only navigation in bronchoscopy (Fig.~\ref{fig:block}). In this section, we detail the pipeline. First, we derive a reduced-order nonholonomic kinematic model of the bronchoscope tip that captures the airway-constrained nature of feasible motions. Second, we introduce a deep learning pipeline for airway lumen detection and temporal tracking, providing geometric landmarks for orientation. Third, we estimate the scope’s forward axis by aggregating vanishing-point cues from detected lumens and infer insertion velocity from depth-based looming. Finally, these measurements are fused within a high-gain observer that enforces airway-following priors and rejects VO drift, yielding smooth and physiologically consistent trajectory estimates.

\subsection{Reduced-Order Kinematic Model of Bronchoscope}
\label{model}

In bronchoscopic navigation, endoscope motion is inherently constrained by the airway anatomy. The bronchoscope advances through tubular passages that restrict lateral deviation, while bends and bifurcations impose nonholonomic constraints on feasible trajectories. Conventional visual odometry (VO), however, tends to emphasize parallax introduced by exploratory maneuvers, wall contacts, or transient lumen occlusions. These effects appear as large apparent motions in the image and can dominate the VO estimate, even though they do not correspond to true progression along the airway. Our goal is therefore to develop a model that suppresses these spurious parallax cues while enhancing sensitivity to low-parallax motion directed into the airway, thereby providing a more faithful characterization of airway geometry and scope progression.
To address this, we aim to recover a smooth and anatomically plausible camera trajectory by incorporating motion priors that capture airway-constrained behavior. Since the bronchoscope predominantly advances along the airway lumen and cannot deviate substantially beyond its tubular boundaries, we formulate a reduced-order, nonholonomic kinematic model of the bronchoscope tip. This model encodes the admissible motions imposed by both user control inputs and the local airway geometry. Building on this prior, we later design an observer that fuses appearance-based visual cues with these kinematic constraints to produce robust and physiologically consistent odometry estimates.

\begin{figure}[t]
    \centering
        \includegraphics[width=0.85\columnwidth]{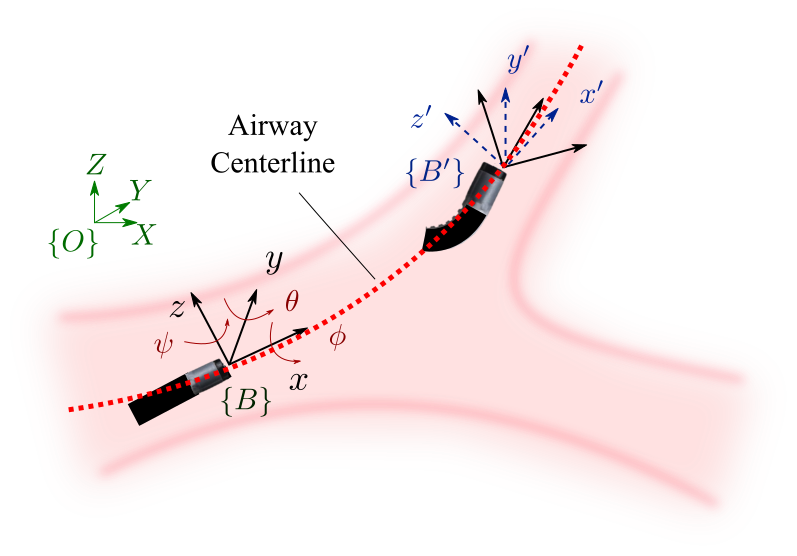}
    \caption{A schematic of reduced-order model of bronchoscopy. An inertial coordinate frame \(O\) is fixed at the entry point of the bronchoscope into the airways at the beginning of the trachea, with the bronchoscope tip position represented by $\mathbf{p}=[ x, \ y, \ z,]^T$.  The operator bends the tip of the bronchoscope ($\theta$) and can twist the bronchoscope to change the plane of bending ($\phi$), allowing navigation through 3D airway structures. A local body-fixed frame \(B\) is attached to the bronchoscope tip, initially coinciding with the inertial frame.
}
\label{fig:schematic}
\end{figure}

Fig.~\ref{fig:schematic} shows a schematic of bronchoscopy. The operator bends the tip of the bronchoscope ($\theta$) and can twist the bronchoscope to change the plane of bending ($\phi$), allowing navigation through 3D airway structures. A local body-fixed frame $\{B\}$ is attached to the tip of the bronchoscope, and let $v$, $\omega$, and $\gamma \in \mathbb{R}$ represent the insertion velocity, twist and bending speed of the bronchoscope, respectively, expressed in the local frame $\{B\}$. Euler angles describe the body frame in 3D space, with roll (\(\phi\)), pitch (\(\theta\)), and yaw (\(\psi\)) corresponding to rotations around the $x$, $y$, and $z$ axes, respectively. To follow the airway centerline, bronchoscope rotation along the $z$-axis is assumed to be negligible ($\psi = 0$). The bronchoscope tip position is represented by $\mathbf{p} = [x, y, z]^T \in \mathbb{R}^3$ relative to the inertial frame. The generalized coordinate of the bronchoscope tip is denoted $\mathbf{q} = [x, y, z, \phi, \theta]^T \in \mathcal{C}$, where $\mathcal{C} \subset \mathbb{R}^5$ is the configuration space.

Under the assumption that the bronchoscope tip follows a perfect path along the $x$-axis, tangent to the airway centerline. Bronchoscope tip rotation around the $y$-axis is a linear function of bending speed, $\dot{\theta} = \gamma$, and internal twist in bronchoscope body is assumed negligible ($\dot{\phi} = \omega$). The rotation matrix $\mathbf{R}_B^O$ for transforming vectors from the local frame to the inertial frame is:

\begin{equation}
\mathbf{R}_B^O = 
\begin{bmatrix}
C_\theta & -S_\theta & 0 \\
C_\phi S_\theta & C_\phi C_\theta & -S_\phi \\
S_\phi S_\theta & S_\phi C_\theta & C_\phi
\end{bmatrix},
\label{eq:rotation_matrix}
\end{equation}
where shorthand notations $S$ and $C$ describe $\sin(\cdot)$ and $\cos(\cdot)$, respectively. Following the assumption that the bronchoscope tip always follows a path along the $x$ axis at a velocity of $v$, we conclude that the bronchoscope's tip's linear velocity is along $y$ and $z$, i.e., $
\mathbf{R}_B^O \ 
[
\dot{x}, \
\dot{y}, \
\dot{z}
]^T
= 
[v, \
0, \
0
]^T
$. Using the rotation matrix $\mathbf{R}_B^O$ in (\ref{eq:rotation_matrix}), we can construct the following constraints:

\begin{subequations}
\begin{align}
- \dot{x} S_\theta + \dot{y} C_\theta C_\phi + \dot{z} C_\theta S_\phi &= 0 \label{eq:constraint1a} \\
- \dot{y} S_\phi + \dot{z} C_\phi &= 0 \label{eq:constraint1b}
\end{align}
\label{eq:constraint1}
\end{subequations}

So far, we have two independent constraints defined by the bronchoscope's kinematics constrained to airway. These constraints are nonholonomic, meaning they cannot be directly integrated or transformed into pose coordinates. They form the basis for the model's reduced-order system. We can write equations (\ref{eq:constraint1}) as a set of Pfaffian constraints \( \mathbf{A}(\mathbf{q}) \dot{\mathbf{q}} = 0 \).

Considering the 5-dimensional configuration space ($\psi = 0$), the system of the two Pfaffian constraints in \eqref{eq:constraint1} entails that the admissible generalized velocities at each configuration \(\mathbf{q}\) belong to the 3-dimensional null space of matrix \(A(\mathbf{q})\). Denoting by $\{g_1(\mathbf{q}), g_2(\mathbf{q}), g_3(\mathbf{q})\}$ a basis of the null space \(\mathcal{N}(A(\mathbf{q}))\), the admissible trajectories for the bronchoscope tip can be characterized as the solution of \( \sum_{i=1}^3 g_i(\mathbf{q}) u_i \), where \(u_i\) are the input vectors for the three remaining degrees of freedom (DoF). The bases of \(\mathcal{N}(A(\mathbf{q}))\) can be calculated from \eqref{eq:constraint1} as follows:

\begin{equation}
\dot{\mathbf{q}} = \begin{bmatrix}
    \dot{x} \\
    \dot{y} \\
    \dot{z} \\
    \dot{\phi} \\
    \dot{\theta} 
\end{bmatrix}
=
\begin{bmatrix}
    \cos(\theta)  \\
    \sin(\theta) \cos(\phi)  \\
    \sin(\theta) \sin(\phi) \\
    0  \\
    0 
\end{bmatrix} v
+\begin{bmatrix}
     0 \\
     0 \\
     0 \\
     1 \\
     0
\end{bmatrix} \omega + \begin{bmatrix}
    0 \\
     0 \\
     0 \\
    0  \\
    1 
\end{bmatrix} \gamma 
\label{eq:kinematic_model}
\end{equation}

\subsection{Airway Lumen Detection and Temporal Tracking}
\label{yolo}
Accurate identification of airway lumens is a critical step for navigation and downstream tasks such as mapping and localization. We developed a deep learning-based detector using the YOLOv8 framework \cite{yolov8_ultralytics}, trained to detect airway lumens from bronchoscopic images. Our training dataset was collected from bronchoscopic procedures on five ex-vivo mechanically ventilated human lungs. A total of 5,000 frames were manually annotated by an expert pulmonologist, who delineated airway lumens using bounding boxes. 

To improve generalization beyond this curated dataset, we adopted a two-stage training pipeline. First, the model was initialized with COCO-pretrained weights and trained on the expert-annotated dataset of 5000 real bronchoscopy images. Second, we employed a semi-supervised bootstrapping strategy by generating 30,000 pseudo-labels from unannotated frames using the initial model, followed by fine-tuning on a weighted mixture of real and pseudo-labeled data. This strategy allowed the detector to leverage a substantially larger dataset while retaining fidelity to expert annotations, resulting in improved robustness across diverse airway appearances.

For temporal consistency across video sequences, we coupled the detector with the SORT algorithm \cite{Bewley2016sort}. SORT (Simple Online and Realtime Tracking) applies a lightweight Kalman filter-based motion model and frame-to-frame bounding box association via the Hungarian algorithm. By integrating YOLO-based airway lumen detection with SORT, our pipeline provides temporally stable lumen tracks with consistent IDs across frames, which is essential for downstream tasks such as camera heading estimation based on vanishing points.

\subsection{Vanishing-Point-Based Orientation Estimation}

In endoluminal navigation, when the bronchoscope advances through the airways the visible lumen openings align along the forward viewing direction of the camera. This phenomenon is naturally described by the concept of vanishing points: straight lines in 3D that are parallel in the world converge in the image plane toward a common point, which defines the direction of camera motion. In the airway setting, the entrances of visible lumens act as approximate portals aligned with the true bronchial direction. Thus, estimating the global vanishing direction from multiple detected lumens provides a geometrically grounded estimate of the scope’s forward orientation. This is particularly valuable because visual odometry inside tubular structures suffers from scale ambiguity and rotational drift; by anchoring motion estimation to lumen-derived vanishing directions, we can obtain stable orientation priors for subsequent odometry.

To this end, we employ the trained YOLO detector in combination with an off-the-shelf monocular depth estimation network~\cite{ZhaFra_BREADepth_MICCAI2025}. Given an input bronchoscopic image $I$, the depth network produces a dense depth map $D(u,v)$, while the YOLO model identifies candidate lumen regions by returning bounding boxes
\begin{equation}
B_i = [x_{1i}, y_{1i}, x_{2i}, y_{2i}], \qquad i=1,\ldots,N ,
\end{equation}
where $(x_{1i}, y_{1i})$ and $(x_{2i}, y_{2i})$ denote the top-left and bottom-right corners of the $i$-th detection. Since the true airway center aligns with the deepest region of each bounding box, we refine the lumen center using depth cues. For each ROI, we extract the corresponding depth submap and compute a percentile threshold (set at the 80th percentile in our experiments). Pixels deeper than this threshold are retained to form the set of candidate lumen pixels within the box. This ensures that only the deepest regions inside each bounding box contribute to the estimation of the lumen center. The refined lumen center is then the centroid of these pixels,
\begin{equation}
(u_i, v_i) = \frac{1}{|\Omega_i|} \sum_{(u,v)\in \Omega_i} (u,v),
\end{equation} 
where $\Omega_i$ is defined as the set of deep pixels inside the $i$-th bounding box that exceed the depth threshold.

Each refined center is back-projected to a normalized camera ray using the intrinsic calibration matrix $K \in \mathbb{R}^{3\times 3}$. The homogeneous image point is
\begin{equation}
\mathbf{x}_i = [u_i,\,v_i,\,1]^\top ,
\end{equation}
which defines a ray in camera coordinates as
\begin{equation}
\mathbf{r}_i = \frac{K^{-1}\mathbf{x}_i}{\|K^{-1}\mathbf{x}_i\|}.
\end{equation}
To emphasize closer lumens, each ray is weighted inversely to the median depth within its ROI,
\begin{equation}
w_i = \frac{1}{\epsilon + \operatorname{median}(D_i)}, \qquad \epsilon \ll 1.
\end{equation}
The global vanishing direction is then estimated by weighted averaging,
\begin{equation}
\mathbf{d} = \frac{\sum_{i=1}^N w_i \mathbf{r}_i}{\left\|\sum_{i=1}^N w_i \mathbf{r}_i\right\|},
\end{equation}
yielding a unit vector $\mathbf{d} = [d_x,\, d_y,\, d_z]^\top$ that points along the estimated scope axis. Orientation angles are finally derived as
\begin{align}
\theta   &= \arctan2(d_x, d_z), \\
\phi &= \arctan2(d_y, d_z).
\end{align}

To complement orientation, we also estimate insertion velocity using the principle of \emph{looming}, which captures radial image expansion of features as the camera moves forward. Let a feature project to image coordinates $(u,v)$ with radial distance from the image center
\begin{equation}
\rho = \sqrt{(u-c_x)^2 + (v-c_y)^2},
\end{equation}
where $(c_x, c_y)$ is the principal point. If the camera translates along its optical axis with velocity $v_z$, the rate of change of this distance satisfies
\begin{equation}
\dot{\rho} \propto \frac{v_z}{Z},
\end{equation}
where $Z$ is the feature depth. In practice, $\dot{\rho}$ is estimated from optical flow between consecutive frames, and its median across tracked features yields a robust proxy for forward velocity: positive values correspond to insertion (features expanding outward), and negative values to retraction (features contracting inward).

In summary, the inputs to the method are: (i) a bronchoscopic image $I$, from which the depth map $D(u,v)$ and airway lumen bounding boxes are obtained, and (ii) the camera intrinsics $K$, determined through prior calibration. The outputs are: (i) fused vanishing-point orientation angles $(\theta, \phi)$, and (iii) an insertion velocity estimate $\dot{\rho}$ up to a scale. Together, these form a complete pose prior—orientation from vanishing-point geometry and translation magnitude from radial feature expansion that can be integrated into a full odometry pipeline.

\subsection{State Estimation via High-Gain Observer}
\label{sec:observer}
We now design an observer that converts noisy, scale-ambiguous visual cues into a smooth airway-consistent tip trajectory. The available measurements are: (i) a position trace $\mathbf p_m(t)$ from any VO/SLAM backend (LoFTR \cite{sun2021loftr} in our experiments), which may drift or be intermittent; (ii) a unit forward direction $\mathbf d_m(t)\in\mathbb S^2$ derived from vanishing-point geometry (Sec.~\ref{yolo}); and (iii) a scaled insertion-speed cue $\tilde v(t)\approx \kappa v(t)$ from looming. Here the subscript $m$ denotes measured quantities: $\mathbf p_m$ is the raw VO position, $\mathbf d_m$ the measured forward direction, and $\tilde v$ the measured (scaled) insertion speed. The observer estimates the bronchoscope tip state
\begin{equation}
\hat{\mathbf q}(t) = \big[\hat{\mathbf p}^\top,\; \hat\phi,\;\hat\theta\big]^\top,
\label{eq:state_vector}
\end{equation}
together with forward velocity $\hat v(t)$ and the unknown scale $\hat\kappa$.

We adopt a high-gain observer because classical stochastic filters such as the Extended Kalman Filter (EKF) or Unscented Kalman Filter (UKF)  are ill-suited to this setting. The dominant failure mode of VO is systematic drift, not zero-mean noise, while the key priors—tube-following and predominantly forward motion—are more naturally imposed through explicit geometric corrections than through process-noise tuning. A high-gain observer directly constructs algebraic innovations that penalize cross-track deviation and spherical misalignment, rapidly re-aligns the estimate after outliers, and remains lightweight enough for real-time deployment.

To this end, we parametrize the forward axis of camera using (\ref{eq:kinematic_model}) by angles $(\hat\phi,\hat\theta)$, giving the unit vector  
\begin{equation}
\hat{\mathbf d}(\hat\phi,\hat\theta)=
\begin{bmatrix}
\cos\hat\theta\\ \sin\hat\theta\cos\hat\phi\\ \sin\hat\theta\sin\hat\phi
\end{bmatrix},
\label{eq:forward_axis}
\end{equation}
with Jacobian
\begin{equation}
\mathbf J(\hat\phi,\hat\theta)=
\begin{bmatrix}
0 & -\sin\hat\theta\\
-\sin\hat\theta\sin\hat\phi & \cos\hat\theta\cos\hat\phi\\
\sin\hat\theta\cos\hat\phi & \cos\hat\theta\sin\hat\phi
\end{bmatrix},
\label{eq:jacobian}
\end{equation}
so that a small increment $\delta\hat{\boldsymbol\eta}=[\delta\hat\phi,\delta\hat\theta]^\top$ produces the change $\delta\hat{\mathbf d}=\mathbf J\,\delta\hat{\boldsymbol\eta}$. With measurements $(\mathbf p_m,\mathbf d_m,\tilde v)$ and estimates $(\hat{\mathbf p},\hat{\mathbf d},\hat v,\hat\kappa)$, we define the following observation errors
\begin{align}
\mathbf e_p &= \mathbf p_m-\hat{\mathbf p},\quad 
e_\parallel=\hat{\mathbf d}^\top\mathbf e_p,\quad 
\mathbf e_\perp=\mathbf e_p-e_\parallel\hat{\mathbf d}, \nonumber\\
\mathbf e_d &= \hat{\mathbf d}\times\mathbf d_m,\quad
r_v=\tilde v-\hat\kappa\,\hat v .
\label{eq:innovations}
\end{align}

Here $\mathbf e_p$ splits into along-track $e_\parallel$ and cross-track $\mathbf e_\perp$ components; $\mathbf e_d$ is the tangent-space orientation error on $\mathbb S^2$ (its norm equals $\sin$ of the misalignment angle); and $r_v$ is the scaled-speed residual. The continuous-time observer dynamics are
\begin{align}
\dot{\hat{\mathbf p}} &= \hat v\,\hat{\mathbf d}
  + \alpha_p\!\left(\mathbf e_\perp+k_\parallel e_\parallel\hat{\mathbf d}\right), \label{eq:obs_p_ct}\\
\begin{bmatrix}\dot{\hat\phi}\\ \dot{\hat\theta}\end{bmatrix}
  &= \alpha_o\,\mathbf J^\top(\hat\phi,\hat\theta)\,\mathbf e_d, \label{eq:obs_ang_ct}\\
\dot{\hat v} &= \alpha_v\,e_\parallel + k_v\,r_v, \label{eq:obs_v_ct}\\
\dot{\hat\kappa} &= \ell_\kappa\,r_v\,\hat v,\qquad 
\hat\kappa\in[\kappa_{\min},\kappa_{\max}] \;\; \text{(projection)}.
\label{eq:obs_kappa_ct}
\end{align}

Equation~\eqref{eq:obs_p_ct} integrates nominal motion while rejecting lateral drift through $\alpha_p\mathbf e_\perp$ and trimming bias with $k_\parallel e_\parallel\hat{\mathbf d}$. Equation~\eqref{eq:obs_ang_ct} aligns orientation by projecting $\mathbf e_d$ through $\mathbf J^\top$, equivalent to gradient descent on $V_o=1-\hat{\mathbf d}^\top\mathbf d_m$. Equation~\eqref{eq:obs_v_ct} corrects speed using both along-track error and the external cue, while \eqref{eq:obs_kappa_ct} adapts the scale factor, driving $r_v\to0$ under sufficient excitation. For implementation, we discretize with period $\Delta t$ via forward–Euler updates, clamp angle and speed increments, and low-pass filter the error terms to suppress outliers. Initial conditions are derived from the first vanishing-point direction, e.g. $\theta_m=\arccos d_{m,x}$ and $\phi_m=\mathrm{atan2}(d_{m,z},d_{m,y})$. With sufficiently large gains $(\alpha_p,\alpha_o,\alpha_v)$, the observer rejects cross-track drift, aligns orientation, corrects along-track bias, and ensures bounded adaptation of $\hat\kappa$. In effect, it fuses VO/SLAM position, vanishing-point orientation, and looming-based speed into a geometry-consistent trajectory robust to the noise, bias, and singularities of bronchoscopic navigation.

\section{Experimental Results}

\begin{figure*}[t!]
    \centering
     \begin{subfigure}[]{\includegraphics[width=0.55\columnwidth]{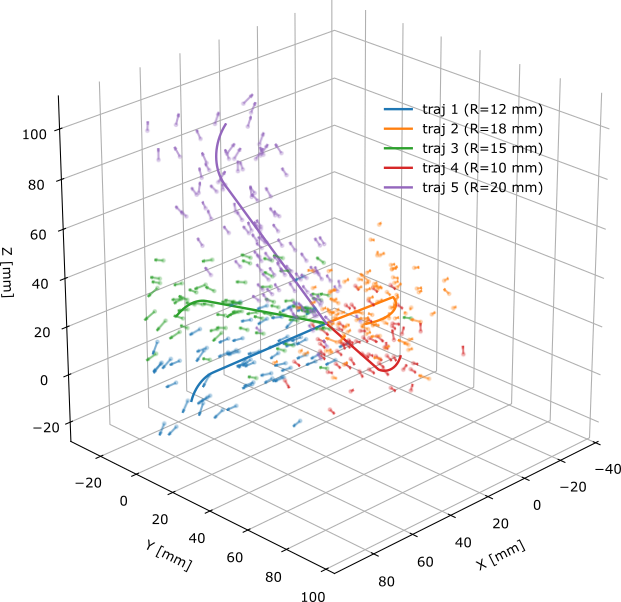} \label{fig:observer_5traj}}
     \end{subfigure}
    \begin{subfigure}[]{
        \centering\includegraphics[width=0.55\columnwidth]{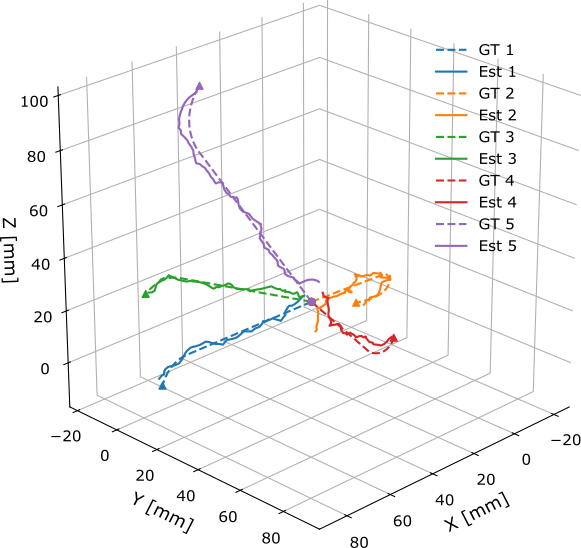} \label{fig:observer_estimate}}
    \end{subfigure}%
    \begin{subfigure}[]{
        \centering \includegraphics[width=0.65\columnwidth]{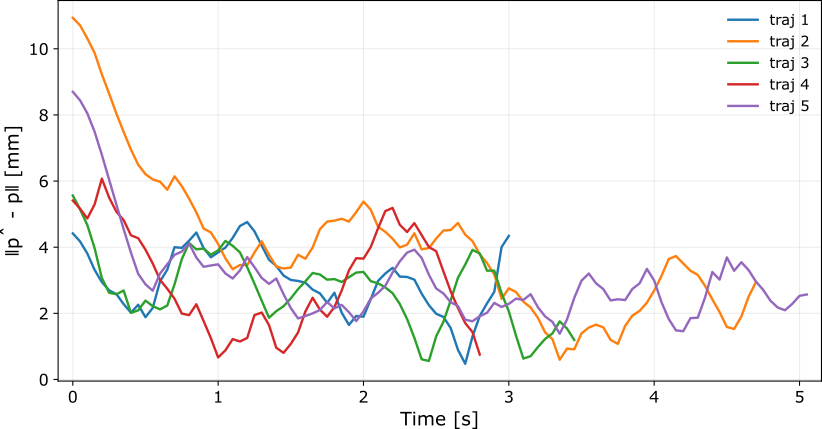} \label{fig:observer_error}}
    \end{subfigure}
\caption{Simulation results for the bronchoscope trajectory observer. 
  (a) A representative sequence showing the airway-constrained trajectory, 
  perturbed measurements, and measured heading arrows. 
  (b) comparison of observer estimates against ground truth for representative trajectories. The observer fuses noisy position/heading to recover an airway-constrained path.
  (c) Instantaneous position error $\|\hat{\mathbf p}-\mathbf p\|$ over time.}
\label{fig:observer_triad}
\end{figure*}

\begin{figure}[tbh]
    \centering
        \includegraphics[width=0.85\columnwidth]{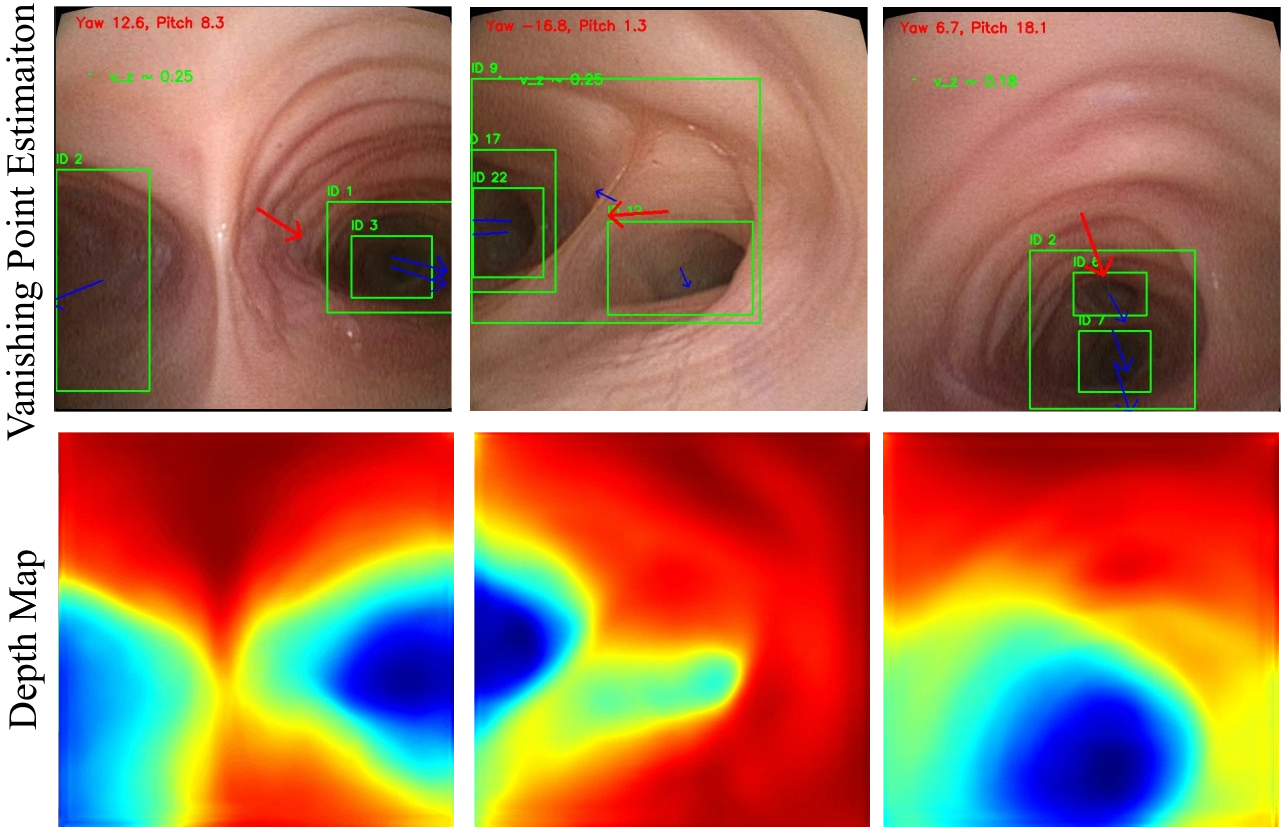}
    \caption{RGB bronchoscopy frames with corresponding depth maps. Airway lumens are detected by YOLO and temporally tracked with SORT (green boxes with IDs). Blue arrows indicate per-lumen rays obtained from back-projection of lumen centers, while the red arrow marks the global vanishing-point heading estimated by weighted fusion of these rays.}
\label{fig:rgbdepth}
\end{figure}

We evaluated the observer in simulation to fine-tune its parameters via trial-and-error. Ten synthetic bronchoscope trajectories were generated, each composed of a straight segment followed by a tangent circular arc (radius $10$–$20$,mm) to mimic an airway bend. All paths start at the origin, are sampled at $\Delta t=50$,ms, and are traversed at constant speed. Measurements were corrupted by i.i.d.\ Gaussian position noise (typ.\ $\sigma_p=20$,mm) and random heading rotations (typ.\ $10^\circ$–$50^\circ$). An auxiliary scaled speed cue $\tilde v \approx \kappa v$ was provided with additive noise, and the unknown scale $\kappa$ was estimated online. The observer fuses $(\mathbf p_m,\mathbf d_m,\tilde v)$ to estimate position $\hat{\mathbf p}$, orientation $(\hat\phi,\hat\theta)$, speed $\hat v$, and scale $\hat\kappa$. In Fig.~\ref{fig:observer_triad} we report five exemplary trials, showing ground-truth vs.\ estimated 3D paths, instantaneous position error $|\hat{\mathbf p}-\mathbf p|$ over time, and per-trajectory RMSE. Errors settle quickly after the initial transient and remain below $\sim6$,mm.

We fixed the observer gains to minimize RMSE error as follows: orientation gain $\alpha_o=15$, position (cross-track) gain $\alpha_p=8$, along-track blend $k_{\parallel}=0.35$, speed channel gain $\alpha_v=6$, speed-cue weight $k_v=1.2$, and scale-adaptation gain $\ell_\kappa=0.2$. These settings yielded fast convergence, smooth estimates, and low steady-state error across all simulated trajectories.

Next, we report the results for the airway lumen detection algorithm in Section~\ref{yolo}. The best-performing model achieved a mean average precision of 0.95 (mAP@0.5) and 0.76 (mAP@0.5–0.95) on the held-out validation set, with precision and recall both around 0.90. These results indicate that the detector is both accurate and balanced, with low false positives and false negatives. Example outputs showing lumen detection, tracking, depth maps, and heading estimation are provided in Fig.~\ref{fig:rgbdepth}.

We validated the proposed framework on an ex-vivo mechanically ventilated human lung connected to a commercial bronchoscope mounted on a 6-DoF Igus ReBeL robotic arm. The robotic arm provided Cartesian insertion and retraction under joystick teleoperation, while a custom actuation unit at the arm’s end-effector offered 2-DoF control of distal bronchoscope bending ($\theta$) and rotation ($\phi$). The lung  was placed inside the magnetic field of an Aurora (NDI) electromagnetic tracking (EMT) system, and a miniature EMT sensor was inserted into the bronchoscope working channel to obtain ground-truth tip pose.  

Eight trajectories were executed, spanning all lobes of the lung, with run durations ranging from $\sim$30\,s to 5\,min. Each sequence contained multiple bifurcations and distal airway insertions, representing clinically relevant navigation challenges. Our algorithm was run on the collected videos, with baseline VO obtained using a LoFTR-based frame-to-frame pipeline. Observer gains were fixed to the values optimized in simulation (Sec.~\ref{sec:observer}) to avoid overfitting. Results were compared against EMT ground truth using Absolute Trajectory Error (ATE) after Sim(3) alignment and Relative Pose Error (RPE) over $\Delta t=10$ frames.  

\begin{figure}[t]
    \centering
        \includegraphics[width=0.9\columnwidth]{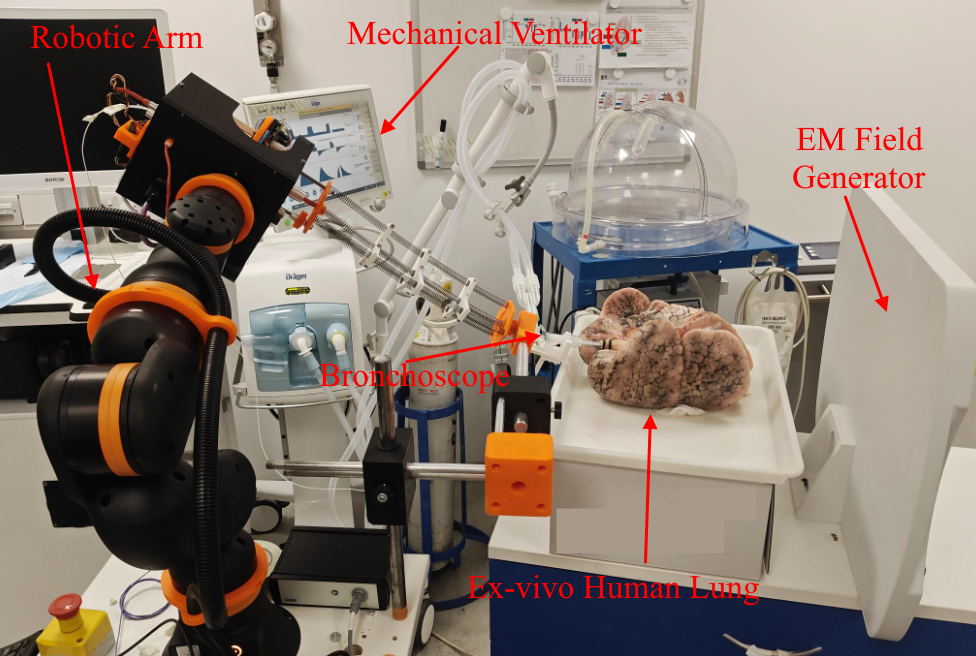}
    \caption{Experimental setup for bronchoscopy validation. 
A commercial bronchoscope is mounted on a 6-DoF Igus ReBeL robotic arm, which inserts and retracts the scope along Cartesian paths into an ex-vivo mechanically ventilated human lung. Joystick commands drive arm motion, while a custom actuation unit at the end-effector provides 2-DoF control of bronchoscope bending ($\theta$) and rotation ($\phi$). The lung is placed within the magnetic field of an Aurora (NDI) EMT system, with a miniature EM sensor inserted at the bronchoscope tip to provide ground-truth pose for evaluation. A total of eight trajectories were executed across all lobes, with lengths ranging from 30\,s to 5\,min.}
    \label{fig:exp_setup}
\end{figure}

\begin{figure*}[t]
    \centering
        \includegraphics[width=1.8\columnwidth]{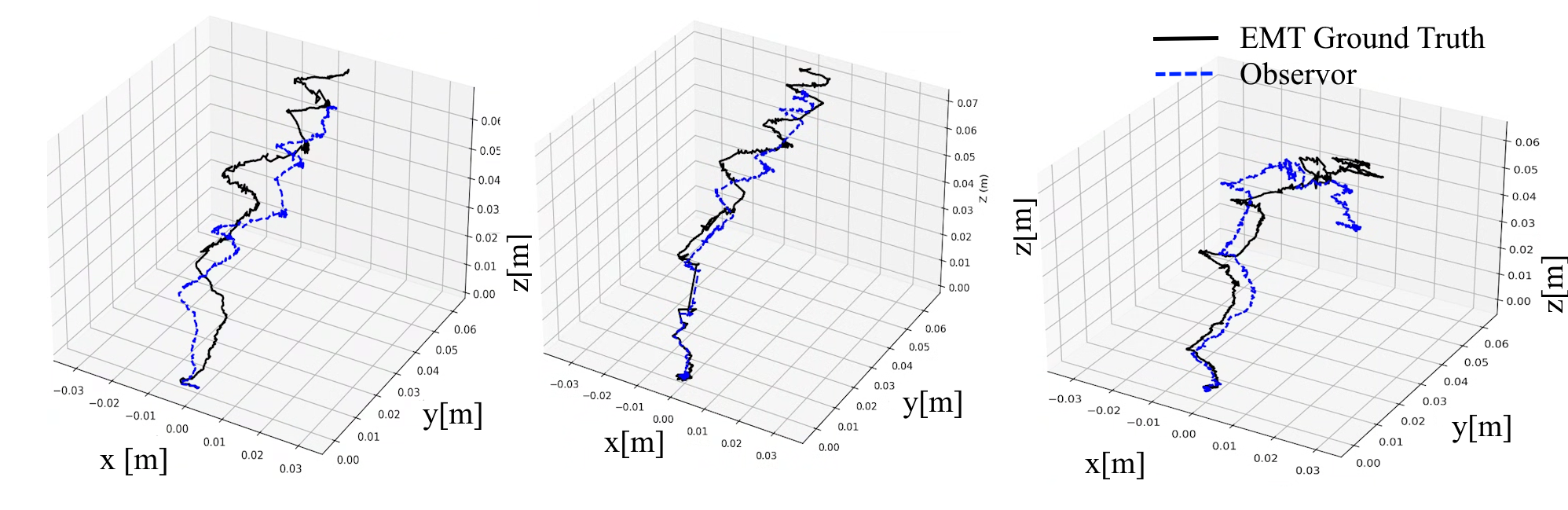}
    \caption{Exemplary results from three trajectories showing observer-based bronchoscope pose estimation compared with EMT ground truth after Sim(3) alignment. The proposed observer achieves close agreement with the reference, demonstrating geometry-consistent tracking across diverse airway paths.}
    \label{fig:observer_results}
\end{figure*}

We evaluated our proposed framework against electromagnetic tracking (EMT) ground truth using both Absolute Trajectory Error (ATE) and Relative Pose Error (RPE), which are the standard benchmarks for odometry evaluation~\cite{sturm2012benchmark}. Since monocular visual odometry (VO) inherently lacks scale, all trajectories were aligned to the ground truth using a full similarity (Sim(3)) transformation via the Umeyama method prior to ATE computation. RPE was computed over short segments ($60$ frames at $10$ fps) to assess local consistency, which is invariant to scale. Exemplary trajectories comparing observer-estimated pose with EMT ground truth after Sim(3) alignment are shown in Fig.~\ref{fig:observer_results}, demonstrating close agreement between estimated and true trajectories across diverse airway paths.  

We compared our method against three state-of-the-art VO pipelines: (i) ORB-SLAM2 (monocular)~\cite{mur2017orb}, a classical feature-based method; (ii) LoFTR-based VO~\cite{sun2021loftr}, which integrates deep feature matching into a frame-to-frame VO pipeline; and (iii) DPVO~\cite{teed2023dpvo}, a recent dense direct monocular VO framework. All baselines were run with default hyperparameters, and sequences were reset on catastrophic tracking failure to ensure a fair comparison.  

\begin{table}[t] \centering \caption{Comparison of VO methods against EMT ground truth. ATE reported after Sim(3) alignment (mm), RPE reported over $\Delta t=10$ frames (deg). Lower is better.} \label{tab:results_vo} \begin{tabular}{lcc} \hline \textbf{Method} & \textbf{ATE (mm)} & \textbf{RPE (deg)} \\ \hline ORB-SLAM2 (mono)~\cite{mur2017orb} & 25.8 & 27.9 \\ LoFTR-VO~\cite{sun2021loftr} & 18.2 & 15.4 \\ DPVO~\cite{teed2023dpvo} & 16.5 & 25.7 \\ \textbf{Ours (Observer Fusion)} & \textbf{11.3} & \textbf{8.6} \\ \hline \end{tabular} \end{table}

Table~\ref{tab:results_vo} summarizes the results. Our approach consistently outperformed all baselines, achieving the lowest ATE and RPE across test sequences. ORB-SLAM2 frequently lost tracking in texture-poor bronchial regions, while LoFTR-VO was more robust but still accumulated significant drift. DPVO performed competitively in well-textured segments but degraded rapidly when parallax cues were absent, which is characteristic of tubular airways. In contrast, our geometry-aware framework, by fusing vanishing-point cues with noisy VO inside a high-gain observer, maintained alignment with airway anatomy and reduced ATE by more than 50\% relative to the best baseline.

\section{CONCLUDING REMARKS}

In this work, we introduced a geometry-aware visual odometry framework tailored to the unique challenges of bronchoscopic navigation. By exploiting vanishing-point geometry to recover heading in parallax-poor regions and fusing these cues with noisy VO through a high-gain observer, we achieved robust, airway-consistent pose estimation. Validation against electromagnetic ground truth in ex-vivo lungs demonstrated that our method substantially outperforms state-of-the-art VO pipelines, reducing both global drift (ATE) and local motion error (RPE).

Several promising directions remain for future work. First, extending the observer to incorporate additional sensing modalities, such as shape sensing or electromagnetic tracking (EMT), could further improve robustness in distal airways while ensuring compatibility with cancer diagnosis pathways. Second, coupling the proposed VO framework with downstream tasks—such as airway mapping, biopsy targeting, or robotic bronchoscopy—has the potential to support fully autonomous or semi-autonomous navigation in critical care. Collectively, these directions aim to bridge the gap between research prototypes and deployable, sensor-free navigation systems for pulmonary interventions.

\bibliographystyle{IEEEtran}

\bibliography{sample.bib}
\end{document}